# Universal Spam Detection using Transfer Learning of BERT Model

Vijay Srinivas Tida
Center of Advanced Computer Studies
School of Computing and Informatics
University of Louisiana at Lafayette
vijay-srinivas.tida1@louisiana.edu

Sonya Hsu, Ph.D.
Informatics Program
School of Computing and Informatics
University of Louisiana at Lafayette
sonyahsu@louisiana.edu

**Abstract**

*Deep learning transformer models become important by training on text data based on self-attention mechanisms. This manuscript demonstrated a novel universal spam detection model using pre-trained Google's Bidirectional Encoder Representations from Transformers (BERT) base uncased models with four datasets by efficiently classifying ham or spam emails in real-time scenarios.*

*Different methods for Enron, Spamassain, Lingspam, and Spamtext message classification datasets, were used to train models individually in which a single model was obtained with acceptable performance on four datasets. The Universal Spam Detection Model (USDM) was trained with four datasets and leveraged hyperparameters from each model. The combined model was finetuned with the same hyperparameters from these four models separately. When each model using its corresponding dataset, an F1-score is at and above 0.9 in individual models. An overall accuracy reached 97%, with an F1 score of 0.96. Research results and implications were discussed.*

## 1. Introduction

Spam email is defined as unsolicited messages in the bulk from Blanzieri and Bryl [1] and constitutes 53.95 percent in the year 2020 [2]. Spam messages create problems because of the internet's availability to most people worldwide. The widespread use of email in many companies results from the quick distribution of information to many people quickly and easily accessible. Spammers access devices in phishing attacks by enticing users to click on the spam link [3]. Another type of threat is email spoofing, in which users believe that it is sent by the person they know [4]. Spam detection tools and techniques are developed by companies so that users will get a better experience. Google's Gmail is one of the largest mail networks claiming 99.9% success of their spam filtering technique [5][6]. Machine learning and deep learning techniques help identify spam emails automatically through trained models [7].

A new entrant to spam processing, transformers showcased their deep learning impacts for Natural Language Processing (NLP) applications [8]. Significant time reduction was achieved due to training the models for better efficiency. Previous methods like Recurrent Neural Networks, Long/Short Term Memory (LSTM) [9], and Gated Recurrent Units (GRU) [10] need to wait for the previous time step information. The data is processed sequentially as the model progressed and challenged to capture long-range dependencies without considering the previously used data points. Instead, transformers parallelize the computation and embed word position with position encoding. Using a multiheaded self-attention mechanism solved the inputs and long-range dependencies [11]. Later, transformers have been proposed as the base of many pre-trained models. Google's BERT (Bidirectional Encoder Representations from Transformers) model has become more popular because it shows higher efficiency in real-world applications and a more straightforward structure. Transformers usually have encoder-decoder architecture whereas, BERT uses only the encoder part – discarding the decoder part. BERT models were trained on the huge amount of data from Book Corpus and English Wikipedia [12]. These models usually produce two outputs. The first output is used for language translation applications like name entity recognition and speech tagging. The other output is used for classification applications like sentiment analysis and fake news detection.

This research aimed to build an efficient universal spam detection model using the pre-trained BERT base uncased model. In this work, the researchers use the second output from the pre-trained Google's BERT base uncased model to find whether the given mail is spam or not since it is a classification problem with the help of the Hugging Face Transformers library [13]. The takeaways from this work are three-fold:

By adding the various layers over the output vector of length 768, the researchers analyzed how to find the sequence length, learning rate, and model architecture for each dataset. Firstly, the pre-trained BERT uncased base model has trained four datasets separately. Secondly, the preprocessing of the hyperparameters, i.e., sequence length and learning rate, speed up the final model architecture selection and training process using the individual trained model. Thirdly, the proposed universal model was computed with recall, precision, F1-score, and accuracy to evaluate different testing sizes and mini-batch sizes from five datasets. The proposed model is the first approach to detecting spam messages using multiple datasets to train the single model.

The article flows as follows: Section 2 illustrated the previous research, whereas Section 3 explained the methodology with detailed descriptions of datasets and modeling the universal model. Section 4 illustrated the final model results, and Section 5 concluded the manuscript with results discussion, research implications, and future research.

## 2. Literature review

Many researchers have shown their effort in spam detection mechanisms. This literature review classified them as machine learning, deep learning, and some combine-based/other approaches. Machine learning and deep learning are approaches to solving real-world problems like image classification and language processing. Machine learning approaches perform well on a small amount of data, whereas deep learning approaches require massive data to surpass the performance of machine learning approaches [13][14]. These two were addressed briefly in the following sections.

## 2.1. Machine Learning-based approaches

Machine learning means designing a machine to learn how to solve a particular task like regression and classification. Machine Learning is given rules or instructions of algorithms to extract features from the data to solve the given task. In machine learning, the programmer should extract features manually.

Harisinghaney [16] tried to implement text and image-based spam emails with the help of the k-nearest neighbor (KNN) algorithm, Naïve Bayes, and reverse DBSCAN algorithms. Preprocessing step of the Enron dataset is done to extract email text before applying any algorithmic approaches to the data using specific feature extraction techniques like Tesseract. The final performance is reported with the help of four metrics precision, sensitivity, and accuracy. Results indicated that all algorithms performed notable well. Youn [17] proposed an ontology-based spam filtering approach and applied the J48 decision tree on the UCI email dataset. The extended work can be seen in [18] Bahgat proposed SVM classifier based on semantic feature selection on the Enron dataset, which showed 94% accuracy. WordNet ontology with some semantic-based methods, principal component analysis, and correlation feature selection methods was used to reduce the number of features to the maximum extent. However, results indicated that logistic regression performed very well. Laorden [19] developed a Word Sense Disambiguation preprocessing step before applying machine learning algorithms to detect spam data. Finally, results indicate a 2 to 6% increase in the precision score when applied on Ling Spam and TREC datasets. George [20] showed KNN based approach achieves higher accuracy when compared to Feed Forward Neural Network. Khonji [21] made an extension for Lexical URL Analysis with the random forest algorithm's help to apply spam detection by their dataset. Jáñez-Martin [22] made the combined model of TF-IDF and SVM showed 95.39% F1-score and the fastest spam classification achieved with the help of the TF-IDF and NB approach. Alberto [23] explained deception detection using various machine learning algorithms with the help of neural networks, random forests, etc.. and paved a path for a new research direction.

## 2.2. Deep Learning-based approaches

Deep learning mimics the human brain to solve the given task without human intervention [24]. Deep Learning uses a neural network with multi-layers with many parameters. In deep learning, automatic extraction of features is accomplished by giving the architecture shape with some hyperparameters. Presently, massive data sets are becoming available to support deep learning approaches. As a result, deep learning techniques show promising results compared to machine learning counterparts in most aspects.

Faris [25] used proposed Feed Forward Neural Network on Spam Assassin dataset in which the Krill Herd algorithm is used for feature extraction. Results indicated that using Krill Herd Algorithm for feature extraction showed better results when compared to other popular training algorithms like the backpropagation and the Genetic based. Raj [26] used the Long Short Term Memory approach and showed an accuracy of about 97% over the Lingspam dataset. Rahman [27] used Bidirectional Long Short Term Memory and achieved 98% accuracy on Lingspam and SPMDC datasets based on their separate implementations. Jie [28] discussed the unsupervised deep learning approach, which can be used for fake content detection in social media.

## 2.3. Combine based / other approaches

Faris [29] proposed a PSO-based Wrapper with a Random Forest algorithm that effectively detects spam messages. Ajaz [30] used a secure hash algorithm with the Naïve Bayes feature extraction method for spam filtering. Van Wanrooij [31] used an IP-based approach and showed their implementations with a better false-positive rate. Lin's [32] system identified spamming botnets with the Bloom filter, which yielded higher precision and values. Esquivel [33] developed IP reputation lists and constantly updated them to perform better than existing models. Regex [34] automatically detected spam/ham mails using regular expressions. This work showed significant performance with minimal computing resources. Marie-Saint [35] introduced the firefly algorithm with SVM and worked with Arabic text. This article concluded the proposed method outperformed SVM alone. Later in [36], Natarajan proposed Enhanced Cuckoo Search for bloom filter optimization. Results showed that ECS outperformed normal Cuckoo search.

Until now, researchers showed significant efforts to detect spam emails with the help of machine learning, deep learning, and combined approaches with the help of algorithms. Most of them tend to design their models based on a specific dataset that might cause the problem in real-world applications. Cross learning of data is necessary to achieve real-time efficiency through advanced deep learning models like transformers and techniques like transfer learning approaches for Natural Language Processing applications. In this work, the researchers solved this issue using four publicly available datasets and showed significant performance in detecting spam messages. This work is carried by designing the model using the output vector from Google's BERT model individually from four datasets with acceptable performance. Then all datasets are combined followed the same process, which showed better performance. The basic terminology is provided in the next section for a better understanding of the methodology and modeling.

Jie [37] discussed the bilingual language multi-type spam detection model using M-BERT, which used image-based spam detection and achieved an accuracy of about 96%. Lee's research [38] showed an accuracy rate of 87% using the Sophos AI proposed CATBERT model by collecting phishing emails. The BERT model was also used for other applications like fake news detection, lie detector, sentiment analysis. Jie [39] also discussed unsupervised deep learning, which is suggested for fake content detection in social media. Further, Barsever [40] proposed a model with the new generative adversarial network to detect lies. In Man's research paper [41], he proposed a sentiment analysis algorithm based on BERT and Convolutional Neural Network with an accuracy rate of 90.5% and 85.2%, respectively.

### 2.3.1. Transformers

Sequential computation load reduction has been a major problem for NLP applications over a long time [42]. NLP is still burdened by linear or logarithmic dependency despite many proposed solutions as the sequence grows [43] [44]. Transformers have simpler architecture without having any convolutional [45] and recurrent layers [9]. The change in architecture solved the problem to a constant number of operations with the help of averaging attention-weighted position, which can be considered Multi-Head Attention and positional embeddings[11]. Transformer models outperformed the existing models with less training cost. Many transformers based models have been invented [46][47][48][49][50][12][51][52]. However, BERT and GPT-2 (Generative Pre-trained Transformer) became the most popular models among the released versions [12][51].

### 2.3.2. Self-attention

Self-attention is the mechanism used to determine the interdependence of tokens in the given input sequence. Self-attention encodes a token by taking information from other tokens. It consists of three weight matrices query, key, and value vectors learned during the training process. Multiheaded self-attention is the extension of self-attention, consisting of multiple sets of the query, key, and value vectors built into transformers[11]. However, this entire process will be managed by the transformers library [11]. The number of heads in multiheaded self-attention is set according to the user requirements, which can be considered a hyperparameter.

### 2.3.3. Transfer learning

Transfer Learning is not only knowledge acquired from pre-trained models for a specific application based on user needs. Usually, these pre-trained models begin with big datasets in which the model's weights contain a lot of information. By fine-tuning the pre-trained model and adding some layers over the pre-trained model's output, the researchers use the same weights from the base model.

### 2.3.4. Parameters and hyper-parameters

Model parameters are considered as weights and biases where the programmer has no control over them. Once the model is defined, then the values will be changed accordingly. On the other side, hyperparameters are given by the user according to the need. For models with better accuracy, hyperparameter selection plays a crucial role which can be obtained by tuning the values 3].

### 2.3.5. Activation functions

Activation functions help determine the neural network's output with the help of some non-linear function to the corresponding output of neurons [54]. The Rectified Linear Unit (ReLU) is added to the neuron outputs at hidden layers [55]. The mathematical equation for ReLU is stated in equation 1:

$$f(x) = \max(0, x) \quad (1)$$

where $f(x)$ is the ReLU activation and x is the input to the function.

Another activation function that is used for multi-class classification problems is the softmax function. The mathematical equation of the softmax function can be stated in equation 2 [56]:

$$f(x) = \frac{\exp(x_i)}{\sum_j (\exp(x_j))} \quad (2)$$

Where $f(x)$ is softmax activation, $x_i$ is the input to the function, and j is the number of classes.

Advanced version softmax is Log softmax function which will apply log to the existing softmax function. The mathematical equation can be stated in equation 3 [57]:

$$f(x) = \log\left(\frac{\exp(x_i)}{\sum_j (\exp(x_j))}\right) \quad (3)$$

Where $f(x)$ is log softmax activation, $x_i$ is the input to the function, and j is the number of classes.

### 2.3.6. Loss Function

The loss function is often considered the cost function to evaluate the model performance given weights [58]. Cross entropy loss is usually used for measuring the performance of a classification-based model whose output probabilities lie between 0 and 1. Equation 4 can represent cross-entropy

$$\text{Cross entropy loss} = -(y \log(p) + (1-y)\log(1-p)) \quad (4)$$

## 3. Methodology

### 3.1. Datasets description

This project used four publicly available datasets, and these are processed such that only the content of the samples is used for training the model.

#### 3.1.1. Ling-spam dataset

This dataset [59] consists of 2893 samples separated into two classes 1) 481 spam messages and 2) 2412 ham messages. This dataset can be accessed from the Kaggle website, which was prepared by modifying the Linguist List. The samples in the dataset focused mainly on job postings, software discussion, and research opportunities areas.

#### 3.1.2. Spam text messages dataset

This dataset [60] consists of 5574 samples separated into two classes 1) 724 spam messages and 2) 4,850 ham messages. The samples in this dataset were collected from the mobile phone spam research-related area. This dataset also was accessed from the Kaggle, which was prepared from the UCI Machine Learning Repository.

#### 3.1.3. Enron dataset

This dataset [61] consists of 32,638 emails separated into two classes 1) 16,544 spam mails and 2) 16,094 ham mails. This dataset can be considered as one of the standard benchmarks in spam classification. This dataset covers a large wide of samples from almost all available options.

#### 3.1.4. Spam assassin

This dataset [62] consists of 6047 emails separated into 1) 1897 spam mails and 2) 4150 ham mails. This dataset can also be considered as one of the standard benchmarks in spam classification. This dataset has two classification levels for ham messages, like easy and hard ham messages, and one for spam messages. However, the unified model presents the combinations of these two kinds of ham messages into one group.

It's a supervised learning model so that labels are noted as 1 or 0 based on spam or not. Then after the model is trained under certain conditions. Label encoding is done for all these samples, in which '0' is considered for ham type, and '1' is considered for spam type data. So along with the text data, these encoded labels are used for training and testing the model's performance.

### 3.2. Data preprocessing and Model Selection Process

The selection of a pre-trained model is essential for the task. Data Preprocessing is considered an essential step for any natural language processing task. However, if using pre-trained datasets, some rules need to be followed according to the model in which conditions were trained. BERT model can be used for applications like generate text embeddings, text classification, named entity recognition, and question answered. To classify spam messages, the BERT model is best for this task as it contains many versions. The base model suits the needs of this research, especially for spam detection. This version contains only 12 encoders with 110 Million parameters which are sufficient for our application.

Unlike transformers, BERT uses only an encoder unit, and the decoder part will be discarded as the name suggests, as shown in Fig. 1. Each encoder consists of the same layers as transformers counterpart, namely Self-Attention and Feed Forward Neural Networks, as shown in Fig. 2. Hence, BERT is considered as a language-based model rather than a sequence-to-sequence-based model. Bidirectional means that the input sequence is processed from both directions so that the model can learn from both directions to predict the word in the context with better efficiency. This model was trained on Wikipedia's unlabelled text corpus (2.5 million words) and book corpus (800 million words). The word representations obtained from the intermediate layers through different weights after training will be helpful for our application in detecting the given input sample is spam or ham. At the end of the model, the design the classifier performs better using adding some neural network layers.

The input sequence will directly feed into the tokenizer as there is no need for any preprocessing steps required for the BERT model. But with some preprocessing steps helped to reduce the sequence length selection which is considered as one of the hyperparameters of the model. Tokenizer will handle the input sequence and perform certain operations on the input data. These operations include tokenizing, contextual and positional encoding, padding, adding unique tokens like (CLS), (SEP), and (PAD), and finally converting the tokenized data into integer sequences. (CLS) and (SEP) tokens are placed at the start and end of the sequence, respectively. Tokenizer implementation can be used directly with the help of the Hugging Face Transformers Library. The output of the architecture consists of two parts. The first output will be used for text classification in which the outputs against the (CLS) tokens are considered. This classifier usually consists of a linear layer and log softmax function. The remaining outputs are used for sequence prediction applications like named entity recognition, parts of speech tagging. The outputs against the unique tokens were discarded.

In this work, the researchers used the second approach where the finetuning process is made by using linear layers, dropout layers [63], batch normalization layers [64], Rectified Linear Unit (ReLU) activations, and log softmax activation with Xavier initialization [65] of weights added at the end in the classifier part of the pre-trained model. The main reason for adding the dropout layer is to avoid overfitting, batch normalization is used to reduce internal covariate shift, and Xavier initialization will help converge the designed model faster.

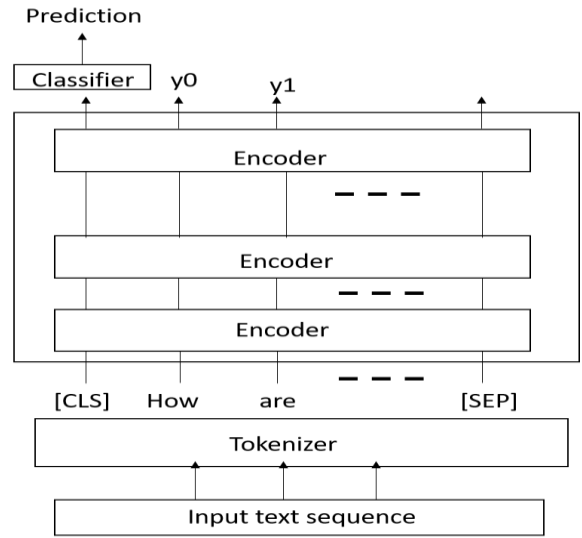

**Figure. 1.** Architecture of BERT

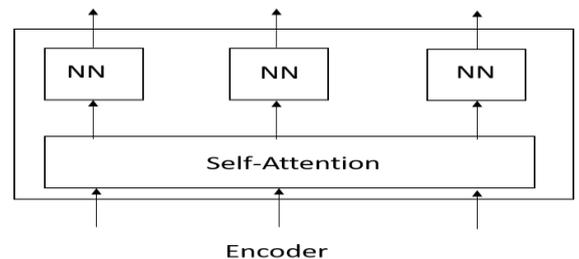

**Figure. 2.** Encoder internal structure [11]

There are two ways to use BERT in our applications. The first approach is to train the model from scratch using the pre-trained weights as initial weights, which requires massive data samples and more computational resources as all the weights are updated after each step. In the second approach, all the pre-trained weights are not updated and require fewer data samples and fewer computational resources. This approach was used.

Hyperparameter tuning is a crucial step for the model to give better performance. Hyperparameters in our proposed model include the sequence length, number of layers, number of neurons, selection of optimizer, learning rate, number of epochs, minibatch size, and selection of layers. The final model should have the best performance by tuning these hyperparameters. This can be done by changing the sequence length, varying the number of neurons in the layers, adding the layers, changing the learning rate of the optimizer until the accuracy of the model is increased. Here Adam optimizer is used for updating the weights in the training process as it has several advantages like computational efficiency and less memory usage with faster training time [66]. In the proposed work, four datasets present different distributions of samples for ham and spam classes. Except for the Enron dataset, all other datasets do not have equal spam and ham samples. Designing model architecture is challenging as the samples from different datasets have different sequence lengths and different numbers of sample distributions. To avoid the problems of biased models, the first models were trained separately on four datasets to analyze the suitable conditions with acceptable performance. Training the standard model from the individual datasets is a crucial step because the Enron dataset has more samples, making the model biased to some extent. While training the individual datasets, it is tough to have the standard architecture for all four datasets to have better performance.

### 3.3. Final modeling

The finalized model was obtained by hyperparameter tuning, which has three fully connected linear layers with batch normalization layers, dropout layers, and some activation functions, which can be seen in Fig .3. In the finalized model, the input was from the output of the [CLS] token side to increase the detection of spam messages. This finalized model has the input of vector length 768, in which this data is passed to the linear layer, which contains 175 neurons. This linear layer will accept 768 as input vector and produces 175 output vector length and hence the shape (768, 175). After this dropout layer of factor 0.1 is placed over the linear layer to ignore 10% of neuron outputs from the linear layer, which will reduce the overfitting of the model. The batch normalization layer is used to make the training faster using reducing generalization error. Activation function ReLU is applied to the output of the batch normalized outputs. Again, the dropout layer is placed at the output of the batch normalization with 0.1 factor.

The combined model performed the best with precision and recall values close to maximum values. The combination minimized the false positives and true negatives. False-positive is considered ham message classified as spam message whereas true negative is considered spam message classified as ham message. Adding the dropout layer before and after the batch normalization layer could avoid the differences of false-positive and true-negative in the trained model on the combined dataset. The dropout layers produced higher precision and recall values with better accuracy and F1-score. Accuracy and F1-score are the metrics for evaluating the model performance. Accuracy is to determine how the model is good in classifying the sampled data to their corresponding classes. The F1-score metric represents the distribution of data samples.

The exact process is repeated one more time, and then a linear layer with shape (100, 2) is placed with log softmax activation to classify whether the input is ham or spam. The output at the end shows if '0', then the input sample is considered ham or if '1', then the input sample is considered spam. After hyperparameter tuning, this finalized model showed better performance compared to other model implementations. The selection of these layers and neurons is based on hyperparameter tuning. Adam optimizer [66] is used here to update the weights as it helps converge the model faster. The learning rate for this optimizer is 3e-4 which showed promising results when compared to other values. The loss function used for this model is a cross-entropy loss.

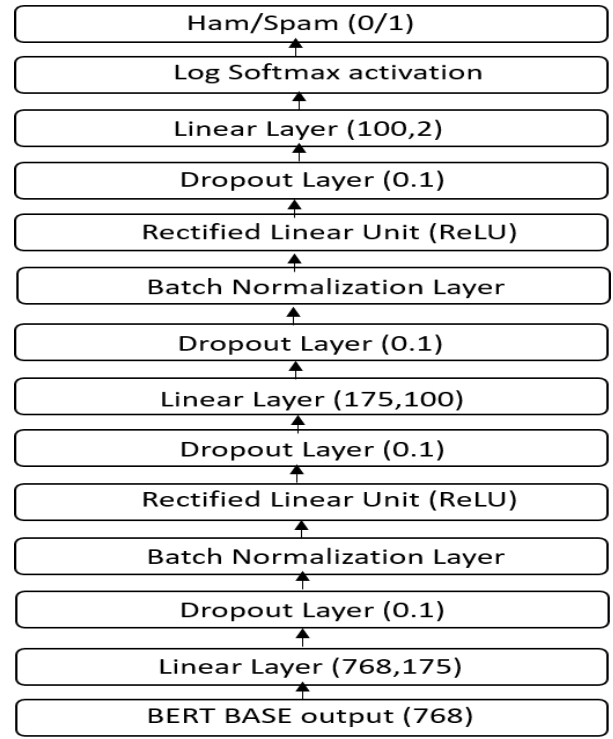

**Figure. 3.** Final model architecture for classifier

When tuning the hyperparameters, the individual models to the same sequence length of the 40 Lingspam dataset model showed a drastic decrease in F1-score to 0.7. If the sample contains words more than sequence length in the pre-trained BERT model, it will be discarded so that the maximum useful content in the sample will be retained by deleting the words in samples that were less than or equal to 3 letters. Since our purpose of using the model is to detect spam messages, we tried to delete the words less or equal to three in their string length which helped in reducing the sequence length. After this preprocessing step, the modified sample is fed as input to the tokenizer. Reducing the sample length of the sample helped improve the F1-score of the Lingspam dataset without affecting much of the other models' performance, which showed an F1-score of above 0.9. After the model architecture is designed with the same hyperparameters for different datasets, the datasets are combined and trained under the same conditions. The finalized model architecture mentioned in Fig. 3 resulted in the highest F1-score at 0.96. This finalized model further changed the hyperparameters with different mini-batch sizes, train-valid-test data distributions, and the number of epochs, which can be further explained in the Hugging Face Transformers library. This library is a collection of transformer implementations that researchers can use directly for their projects. These implementations are compatible with TensorFlow and PyTorch APIs [67]. Google's BERT base uncased model is used to implement this library with PyTorch related to this project. To further improve the performance, gradient clipping was used to prevent the model from exploding gradient problem, that is set to 1.0 [68], and model checking process, which makes the model save the weights corresponding to the less validation loss. The PyTorch framework was used for training the model, which has predefined data loaders to help create batches, shuffle and load the data in parallel using multiprocessing [69][70].

## 4. Results

This section will discuss the performance metrics and results obtained from the finalized model as discussed in section IV by varying minibatch sizes from 16 to 1024, different train-valid-test distributions like 60:20:20, 70:15:15, 80:10:10 and epoch size set to 200.

Different performance metrics are considered while evaluating our proposed model and is explained as below:

**Table1.** Confusion Matrix

| Real Result | Test Result Predicted | |
|---|---|---|
| | HAM | SPAM |
| HAM | TN | FP |
| SPAM | FN | TP |

### 4.1. Accuracy

As shown in table1, the metric will help to visualize the model's performance. Confusion Matrix consists of four elements which are defined as below

a) TN: True Negative in which ham sample predicted as ham.
b) TP: True Positive in which spam sample predicted as spam.
c) FP: False Positive in which spam sample predicted as ham
d) FN: False Negative in which ham sample predicted as spam

Classifying whether the model performed well or not is simply dividing correct predictions by all predictions. Measuring this metric, the sklearn library was used. The formula for defining accuracy is defined below [71]:

$$Accuracy = \frac{(TN+TP)}{(TP+FN+FP+TN)} \quad (5)$$

### 4.2. Recall

Recall measurement is defined as the number of spam samples that are correctly predicted among all the spam samples provided in the dataset. The formula for defining recall is defined in equation 6 [71]:

$$Recall = \frac{TP}{TP+FN} \quad (6)$$

### 4.3. Precision

Precision measurement is defined by the number of samples classified from the given set of positive samples. In other words, how many samples were correctly predicted as spam from the total number of samples predicted positive?

The formula for defining precision is defined in equation 7 [71]:

$$Precision = \frac{TP}{TP+FP} \quad (7)$$

### 4.4. F1-Score

F1- score is defined as the harmonic mean of precision and recall values. The formula for F1-score can be defined in equation 8 [71]:

$$f1\text{-}score = \frac{2\times(precison \times recall)}{(precision+recall)} \quad (8)$$

To analyze the finalized version of the model, the varied batch size, the number of epochs, and train-valid-test data distribution were implemented. Batch size is considered the number of input samples is fed to the model at a time. Epochs are considered the number of times the same samples are repeated and fed to the model during the training phase. Train-valid-test data distribution is considered as dividing the dataset into three parts training, validation, and testing parts from the given dataset. In this work, 60:20:20, 70:15:15, and 80:10:10 were used to train valid and test distributions. The training part is considered for training the model. The validation part is used to visualize whether the model is learning properly or not during the corresponding training phase. The testing part is the final step to analyze the performance of the trained model.

**Table 2.** F1 and acuracy values for the different datasets

| Dataset | Minibatch size | Distribution | Highest f1-score | Accuracy |
|---|---|---|---|---|
| SpamAssassin | 128 | 70:15:15 | 0.9764 | 0.98 |
| Enron | 128 | 70:15:15 | 0.9720 | 0.97 |
| LingSpam | 512 | 80:10:10 | 0.9400 | 0.98 |
| SpamText | 128 | 80:10:10 | 0.9396 | 0.98 |
| Combined | 128 | 70:15:15 | 0.9608 | 0.97 |

**Table 3.** Corresponding precision and recall values

| Dataset | Minibatch size | Distribution | Precision | Recall |
|---|---|---|---|---|
| SpamAssassin | 128 | 70:15:15 | 0.96 | 0.99 |
| Enron | 128 | 70:15:15 | 0.96 | 0.98 |
| LingSpam | 512 | 80:10:10 | 0.90 | 0.98 |
| SpamText | 128 | 80:10:10 | 0.95 | 0.93 |
| Combined | 128 | 70:15:15 | 0.95 | 0.97 |

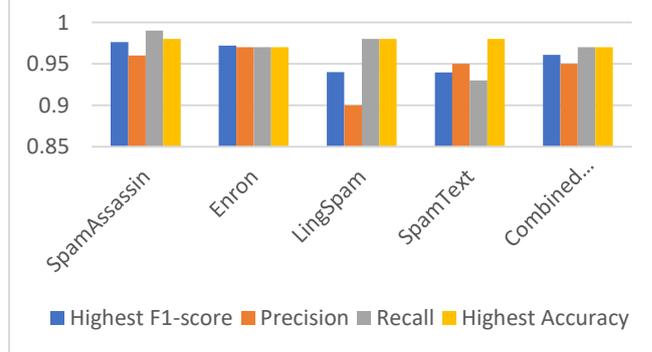

**Figure. 4**. Accuracy: F1 values with precision and recall values

The results of the finalized model using four datasets separately and then, after getting acceptable performance, will perform the same process by combining all four datasets. The number of epochs is set to 200, which helped the model for all four datasets individually showed acceptable performance. After model evaluation, the best case from the individual and combined dataset are shown in Tables 2 and 3. 200 epochs by varying minibatch size from 16 to 1024 was performed. Finally, the highest accuracy, F1-score, and corresponding recall precision values for all the four and combined datasets are visualized from Fig. 4. The results indicated that batch size 128 is the best fit for the models trained on five datasets. The combined dataset achieved 97% accuracy with a 0.96 F1-score by the hyperparameters from the individually trained models. Although the Lingspam dataset showed more accuracy at minibatch size 512, the combined dataset performed better at 128 batch size.

## 5. Conclusion and Future work

Previous models of the past research showed their results based on the individual datasets using several techniques. Specifically, those models trained on individual datasets with accuracy; however, their performance might vary if they were to replicate on other datasets. The researchers intended to address this issue by combining all datasets with training the model for better accuracy in this manuscript. The combined model outperformed those models trained on individual datasets.

In this manuscript, the USDM is provided to solve the different results by individual datasets. It can be helpful in the real-time scenario for spam classification with the USDM using BERT based on the combination of different datasets as an input to the designed model. Based on the individually trained model of multiple datasets and added dropout layers above and below the batch normalization layers, the USDM performed better with an F1 score and acceptable precision and recall values. The designed model achieved a 97% of accuracy at an F1 score of 0.97. With frequently released transformer models, a better model may be used for better spam data detection with less training time.

Another problem worth mentioning is overtrained models in real-time classifications. If the model is trained using a single dataset, it doesn't relate to other samples. By adding more data samples, the model performs better than the overfitted model.

This manuscript can be extended to various applications, i.e., fake news detections from social media platforms, inadequate content filtering from online sources, and such. Deep Learning models have higher accuracy when feeding more data samples. Deep learning research mainly focused on a specific dataset might not have a desirable outcome. This manuscript presented the first attempt at combining datasets using a deep learning approach to provide a better model. Further research in combining multiple datasets is encouraged to further validate the USDM.